%% file: acl_latex.tex
\documentclass[11pt]{article}

\usepackage[final]{acl}

\usepackage{times}
\usepackage{latexsym}

\usepackage[T1]{fontenc}

\usepackage[utf8]{inputenc}

\usepackage{microtype}

\usepackage{inconsolata}

\usepackage{graphicx}

\usepackage{amsmath}
\usepackage{amssymb} 

\usepackage{float}
\usepackage{stfloats}

\usepackage{algorithm}
\usepackage{algpseudocode}

\usepackage{multirow}

\usepackage{pifont}
\newcommand{\cmark}{\ding{52}}
\newcommand{\xmark}{\ding{56}}

\usepackage{booktabs}
\usepackage{array}

\newcommand{\rowdot}[1]{%
  \par\noindent
  \hangindent=1.4em 
  \hangafter=1     
  \makebox[1.4em][l]{\textbullet}#1%
}

\usepackage[table]{xcolor}

\title{UT-ACA: Uncertainty-Triggered Adaptive Context Allocation \\ for Long-Context Inference}

\author{
  Lang Zhou\textsuperscript{1,2}, 
  Shuxuan Li\textsuperscript{1}, 
  Zhuohao Li\textsuperscript{1,2},
  \textbf{Shi Liu\textsuperscript{2,3}}, \\
  \textbf{Wei-Shi Zheng\textsuperscript{1,2}},
  \textbf{Zhilin Zhao\textsuperscript{1,2}\thanks{Corresponding author.}} \\
  \\ 
  \textsuperscript{1}\,School of Computer Science and Engineering, Sun Yat-sen University \\ 
\textsuperscript{2}\,Shenzhen Loop Area Institute \quad
\textsuperscript{3}\,Southern University of Science and Technology \\
  {zhoulang3}@mail2.sysu.edu.cn, zhaozhlin@mail.sysu.edu.cn \\
}

\begin{document}
\maketitle
\begin{abstract}

Long-context inference remains challenging for large language models due to attention dilution and out-of-distribution degradation. Context selection mitigates this limitation by attending to a subset of key-value cache entries, yet most methods allocate a fixed context budget throughout decoding despite highly non-uniform token-level contextual demands. To address this issue, we propose Uncertainty-Triggered Adaptive Context Allocation (UT-ACA), an inference-time framework that dynamically adjusts the context window based on token-wise uncertainty. UT-ACA learns an uncertainty detector that combines semantic embeddings with logit-based confidence while accounting for uncertainty accumulation across decoding steps. When insufficient evidence is indicated, UT-ACA selectively rolls back, expands the context window, and regenerates the token with additional support. Experiments show that UT-ACA substantially reduces average context usage while preserving generation quality in long-context settings. Code and dataset are available at \url{https://github.com/Tommy307/UT-ACA}.
\end{abstract}

\input{chap/1-intro.tex}
\input{chap/2-pre.tex}
\input{chap/3-method.tex}
\input{chap/4-exp.tex}
\input{chap/5-rela.tex}


\section{Conclusion}
We propose UT-ACA, an uncertainty-triggered adaptive context allocation framework that dynamically adjusts the context window at the token level during decoding. Tokens are tentatively generated under a compact window and regenerated with expanded context only when the uncertainty detector signals insufficient evidence, enabling selective allocation without a fixed per-token budget. Experiments across synthetic and out-of-domain long-context benchmarks confirm that UT-ACA substantially reduces average context token consumption while maintaining competitive generation quality.

\section{Limitations}
The proposed UT-ACA significantly reduces the average number of context tokens, while the per-token decoding latency does not decrease proportionately.
We attribute this phenomenon to the computational overhead introduced by the rollback mechanism. Uncertain tokens require regeneration after context expansion, leading to increased inference time, particularly when uncertainty is triggered frequently. Exploring alternative decoding strategies that reduce rollback frequency or amortize its cost remains an important direction for future research.

\section*{Acknowledgments}
This work was partially supported by the NSFC General Program (62576367), the NSFC Excellent Young Scientists Fund (Overseas) (2025HY00260105 and 2025HYSPT0708), and the Fundamental Research Funds for the Central Universities, Sun Yat-sen University (25hytd012).


\bibliography{custom}

\clearpage
\appendix

\input{chap/appendix.tex}


\end{document}

%% file: chap/1-intro.tex
\section{Introduction}
Large language models (LLMs) are expected to operate in long-context settings, such as long-document question answering, multi-document summarization, and evidence-grounded reasoning over extensive inputs~\cite{zhang2025academiceval, wu2025longgenbenchbenchmarkinglongformgeneration, li-etal-2024-loogle}. In these scenarios, effective long-context handling is pivotal for robust generation, requiring reliable evidence retrieval and globally consistent generation. However, as context length increases, redundant or weakly relevant tokens dilute attention and hinder precise evidence retrieval. 
Furthermore, long contexts introduce out-of-distribution (OOD) dependencies and positional patterns that diverge from pretraining data~\cite{ding2024longrope, peng2024yarn, li2024longcontextllmsstrugglelong}, resulting in degraded calibration and unstable reasoning during length extrapolation.

\begin{figure}[t]
  \begin{center}
    \includegraphics[width=\columnwidth]{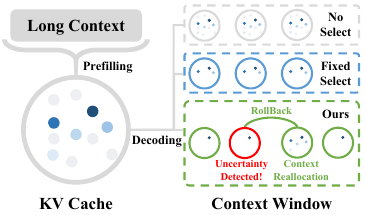}
    \caption{Comparison of context management strategies. The figure contrasts full-context decoding, fixed-size context selection, and the proposed adaptive context allocation strategy. Full-context decoding attends to all available KV entries, fixed selection uses the same retrieved budget at every step, and UT-ACA expands the window only when uncertainty indicates that additional evidence is needed.}
    \label{fig:comparison_with_diff_ways}
  \end{center}
\end{figure}


Context selection has emerged as an effective strategy for mitigating long-context inference challenges by restricting attention to relevant Key-Value (KV) entries, thereby reducing attention dilution and computational overhead. Nevertheless, most existing approaches~\cite{wu-etal-2025-tokenselect, hao2025omnikv, pmlr-v235-tang24l} maintain a fixed context budget throughout decoding, which implicitly assumes that generation difficulty remains uniform across tokens. This assumption is often violated in practice: many tokens can be reliably predicted from local context, whereas others depend on long-range evidence dispersed across the prompt. Consequently, fixed-size context windows may be unnecessarily costly for easy generation steps yet inadequate for difficult ones, highlighting the need for adaptive token-wise context allocation during decoding.


A natural implication is that context requirements vary at the token level and should be adjusted dynamically during decoding. Token-wise generation difficulty can be inferred from uncertainty signals available at inference time, enabling adaptive context control: decoding generally defaults to a compact window, and when high uncertainty or hallucination risk is detected, the process reverts, expands the context window, and regenerates the token with additional evidence. In this formulation, uncertainty estimation serves as an active decision signal for selective context enlargement rather than a passive diagnostic measure.

Accordingly, we propose \textbf{UT-ACA}, an inference-time framework that dynamically adjusts the context window based on token-wise uncertainty. As shown in \figurename~\ref{fig:comparison_with_diff_ways}, UT-ACA generally decodes with a compact context window and expands it only when insufficient evidence or elevated hallucination risk is detected. Uncertainty is estimated via a learned dual-encoder detector combining logit margin and semantic embeddings from hidden states, a lightweight strategy that avoids the overhead of multi-sample decoding~\cite{khairi-etal-2025-life, li2025speculativedecodingmultisampleinference, Kuhnetal2023}. To train the detector, we construct a synthetic dataset of $10{,}000$ training, $120$ validation, and $100$ test samples from LLM output logits and hidden states, and adopt a dual-encoder architecture that fuses semantic and confidence features with an LSTM-based temporal module. Experiments show that UT-ACA generalizes across diverse long-context settings, achieving comparable or improved generation quality while reducing context token consumption.

%% file: chap/2-pre.tex
\section{Preliminaries}

At inference time, LLMs maintain a KV cache of past token representations across 
attention layers~\cite{hooper2024kvquant, liu2024minicache}. At each decoding step $t$, 
we define the \emph{context window} $\mathcal{C}_t \subseteq \{(K_i, V_i)\}_{i=1}^{t-1}$ 
as the subset of KV entries participating in attention computation, and self-attention 
is computed as
\begin{equation}
    \mathrm{Attn}(Q_t, \mathcal{C}_t)
    = \mathrm{softmax}\!\left(
    \frac{Q_t K_{\mathcal{C}_t}^\top}{\sqrt{d}}
    \right) V_{\mathcal{C}_t}.
\end{equation}
In long-context settings, attending to the full KV cache is computationally expensive and 
often unnecessary~\cite{An2024Make}, motivating adaptive strategies that dynamically 
adjust $|\mathcal{C}_t|$.

\paragraph{Rollback and Regeneration.}
Autoregressive generation is susceptible to error accumulation~\cite{huang2024selfcorrect}: 
uncertain or incorrect predictions at earlier steps propagate through the KV cache and 
affect subsequent decoding. We denote the decoding state at step $t$ as $\xi_t = 
(x_{<t}, \mathcal{C}_t)$. When a token is generated under insufficient contextual 
evidence, a rollback operation restores the pre-generation snapshot for step (t),
after which the token is regenerated under an expanded context window.

\paragraph{Problem Formulation.}
Given a long-context input and an autoregressive LLM, this work studies how to adaptively allocate the context window $\mathcal{C}_t$ at each decoding step. Rather than committing to a fixed context budget throughout decoding, the goal is to default to a compact window when local evidence is sufficient, and to detect uncertain steps that warrant rollback, context expansion, and regeneration. In this way, the model aims to reduce average context token consumption while preserving generation quality in long-context inference.

%% file: chap/3-method.tex
\section{UT-ACA Framework}

\begin{figure*}[t]
  \centering
  \includegraphics[width=\textwidth]{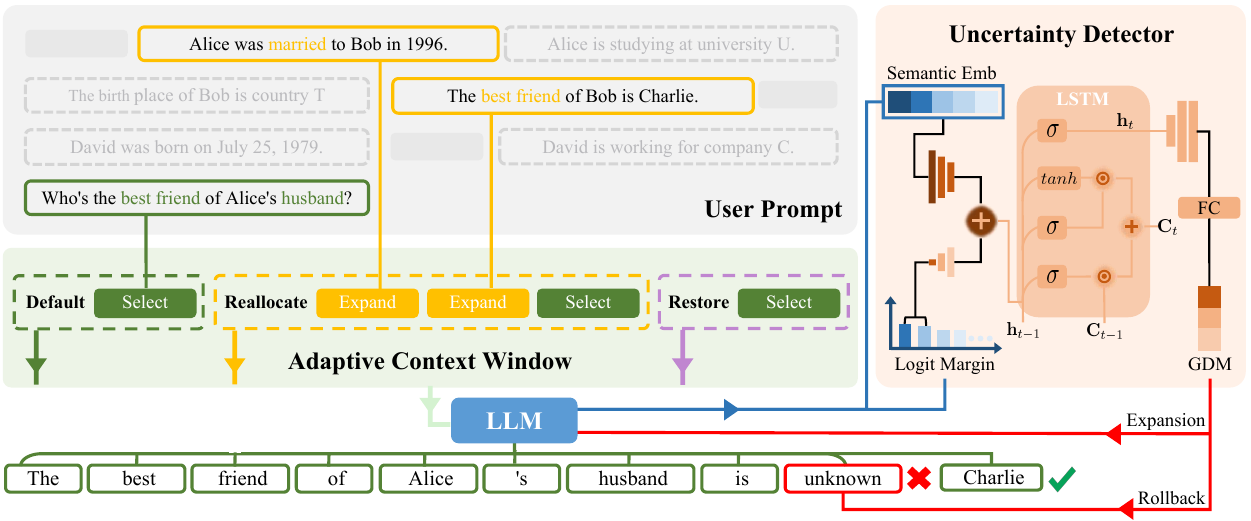}
  \caption{Workflow of our framework. The figure illustrates how UT-ACA connects long-context input, token-level uncertainty estimation, and adaptive context allocation during decoding. (a) The prompt provides the long-context input and target instruction. (b) The uncertainty detector converts output logits and semantic embeddings into a Generation Difficulty Metric (GDM). (c) The adaptive context window accepts low-uncertainty tokens, but rolls back, expands the retrieved context, and regenerates the token when the GDM indicates insufficient support.}
  \label{fig:main_stream}
\end{figure*}

We propose \textbf{UT-ACA}, an inference-time framework consisting of two core modules, the \emph{uncertainty detector} and the \emph{adaptive context window}, which jointly enable token-wise dynamic context allocation during decoding, as illustrated in \figurename~\ref{fig:main_stream}. At most decoding steps, UT-ACA performs a tentative generation with a compact context window and feeds both the output logits and hidden-state embeddings into the \emph{uncertainty detector} to obtain the \emph{Generation Difficulty Metric} (GDM). Conditioned on the GDM, the \emph{adaptive context window} module either accepts the tentative token and proceeds with the compact window, or triggers rollback, context expansion, and regeneration with enriched contextual support.

\subsection{Uncertainty Detector}
\label{subsec:uncertainty-detector}
To support uncertainty-aware decoding, we design a lightweight token-level uncertainty estimator that operates at each generation step. The estimator takes two complementary signals as input, namely the logit margin and a semantic embedding extracted from model hidden states. It then fuses these signals with a dual-encoder module and models uncertainty accumulation over time with a Long Short-Term Memory (LSTM) layer~\cite{Zhao_2025_STEM-LTS, qiu-etal-2024-large}, producing a three-way generation difficulty metric that is later used to trigger context reallocation.

\paragraph{Input signals.} At decoding step $t$, let $\boldsymbol{\ell}_t \in \mathbb{R}^{\mathcal{V}}$ denote the output logits, where $\mathcal{V}$ denotes the size of the LLM vocabulary. We adopt the logit margin $m_t$ as a lightweight confidence signal
\begin{equation}
m_t = \boldsymbol{\ell}_t^{\langle 1\rangle} - \boldsymbol{\ell}_t^{\langle 2\rangle},
\end{equation}
where $\boldsymbol{\ell}_t^{\langle 1\rangle}$ and $\boldsymbol{\ell}_t^{\langle 2\rangle}$ are the largest and second largest logits. Margin alone is not a reliable uncertainty proxy because semantically similar candidates may receive logits with similar values, resulting in a comparatively small logit margin during generation. To handle this issue, we extract a token semantic embedding from the inner states of the LLM. Following prior work on hidden-state analysis~\cite{ma2025estimating, ferrando2024doiknow, singh2024needs, chen2024inside}, 
we use the output of the attention sub-layer in the final transformer layer as $\mathbf{emb}_t$ to represent token-level semantic concepts.

\paragraph{Dual-encoder fusion with temporal modeling.} We fuse the semantic embedding $\mathbf{emb}_t$ and the logit margin $m_t$ with a dual-encoder module, which preserves the complementary roles of semantic complexity and score-based confidence while aligning them in a shared space for joint decision making. Concretely, the two branches map both signals into a $d$-dimensional representation and yield
\begin{equation}
\mathbf{z}_t=f_e(\mathbf{emb}_t)+f_m(m_t),
\end{equation}
where $f_e$ and $f_m$ denote the semantic and logit margin encoders, respectively. Element-wise addition is applied to generate a fused representation $\mathbf{z}_t$.
Since uncertainty can propagate across autoregressive steps and compound after an early mistake, we apply an LSTM to explicitly model this temporal accumulation and aggregate historical evidence
\begin{equation}
\mathbf{h}_t, \mathbf{c}_t = \mathrm{LSTM}\!\left[\mathbf{z}_t, \mathbf{h}_{t-1}, \mathbf{c}_{t-1}\right],
\end{equation}
where the memory state $\mathbf{c}_t$ captures past uncertainty and the hidden state $\mathbf{h}_t$ serves as the step-wise generation difficulty representation for downstream prediction.

\begin{figure}[t]
  \begin{center}
    \includegraphics[width=\columnwidth]{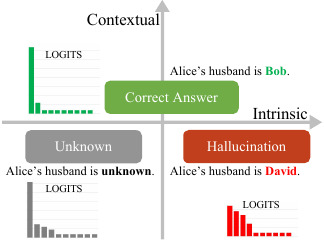}
    \caption{Token-generation scenarios for the Generation Difficulty Metric. The figure categorizes grounded answers, abstentions, and hallucinations by whether the required information is supported by the context or by the intrinsic knowledge of the model.}
    \label{fig:generation_difficulty_metric}
  \end{center}
\end{figure}

\paragraph{Output as generation difficulty metric.} We define the \emph{generation difficulty metric} (GDM) as a three-way probability vector over the token-generation scenarios shown in \figurename~\ref{fig:generation_difficulty_metric}, including grounded correct generation (Correct Answer), unknown-style abstention (Unknown), and unsupported content generation (Hallucination). This three-way design is central to UT-ACA because it separates reliable generation from two distinct failure modes under insufficient contextual evidence. The prediction head maps $\mathbf{h}_t$ to the GDM:
\begin{equation}
\begin{aligned}
& \mathbf{p}_t=\mathrm{softmax}\!\left(g_{\phi}(\mathbf{h}_t)\right), \\
& \mathbf{p}_t = \left[p_{t,\mathrm{hal}},\ p_{t,\mathrm{cor}},\, p_{t,\mathrm{unk}}\right], \\
& \mathbf{p}_t \in \mathbb{R}^{3}, \quad \sum_{k \in [3]} p_{t,k} = 1,
\end{aligned}
\end{equation}
where $g_{\phi}$ denotes the prediction head, and $p_{t,\mathrm{cor}}$, $p_{t,\mathrm{unk}}$, and $p_{t,\mathrm{hal}}$ correspond to correct, unknown-style abstention, and hallucination, respectively.

A binary formulation is insufficient for this setting because it merges \emph{Unknown} and \emph{Hallucination} into a single negative class, despite their different confidence patterns. Unknown tokens often correspond to confident abstention phrases and can exhibit high logit margins comparable to correct predictions, whereas hallucinated tokens introduce unsupported content. This mismatch introduces label noise and weakens the ability of the detector to identify when additional context is needed. In practice, the downstream policy triggers context reallocation whenever the probability mass assigned to non-grounded states becomes dominant:
\begin{equation}
\max(p_{t,\mathrm{hal}}, p_{t,\mathrm{unk}}) > p_{t,\mathrm{cor}},
\end{equation}
which expands the context window and regenerates the current token for rectification.

\subsection{Adaptive Context Window}

\begin{algorithm}[t]
\caption{UT-ACA Framework}
\label{alg:refined_greedy}
\begin{algorithmic}[1]
\Require Model $\mathcal{M}$, Detector $\mathcal{D}$, Input $x_0$
\State Initialize cache $\mathbf{C}$, context window size $k \gets K_\text{max}$, $t \gets 0$

\While{$x_t \neq \text{<EOS>}$}
    \State $\mathcal{S} \gets \text{Snapshot}(\mathbf{C})$ \Comment{Save state}
    
    \State \emph{// Step1: Tentative Generation}
    \State $\hat{x}, \hat{\mathbf{C}} \gets \mathcal{M}(x_t, \mathbf{C}, \text{topk}=k)$
    
    \State \emph{// Step2: Uncertainty Check}
    \State $is\_unsafe \gets \mathcal{D}(\hat{x}, \text{activations})$
    
    \If{$is\_unsafe$ and $k < K_{max}$}
        \State $\mathbf{C} \gets \text{Restore}(\mathcal{S})$ \Comment{Roll back}
        \State $k \gets K_\text{max}$ \Comment{Expand}
        \State \textbf{Regenerate:} \State $x_{t+1}, \mathbf{C} \gets \mathcal{M}(x_t, \mathbf{C}, \text{topk}=k)$
    \Else
        \State $x_{t+1}, \mathbf{C} \gets \hat{x}, \hat{\mathbf{C}}$
        \State $k \gets \text{UpdatePolicy}(k)$ \Comment{Shrink}
    \EndIf
    
    \State $t \gets t + 1$
\EndWhile
\end{algorithmic}
\end{algorithm} 

Inspired by InfLLM~\cite{xiao2024infllm}, we adopt fixed-size blocks as the basic units of our adaptive context window. During prefilling, we partition the KV representations of the long context into blocks that contain the same number of tokens. For each block, we keep a small set of representative tokens to support efficient relevance estimation. During decoding, the query vector at step $t$ retrieves the top-$k$ most relevant blocks, and attention is computed only over the selected blocks to mitigate long-context OOD behaviors.

Since token difficulty varies across steps, the required contextual evidence is non-uniform. UT-ACA initialized at $K_\text{max}$ for the first token to establish contextual grounding and subsequently adjusted by the update policy. At each step, the output logits and hidden-state embeddings are fed into the uncertainty detector to obtain the GDM. When either of the non-grounded probabilities exceeds the grounded probability, UT-ACA rolls back the tentative token, expands the context window by retrieving more relevant blocks, and regenerates the current token. Formally, let $\mathcal{B}$ be the set of all blocks, $\mathbf{r}(B)$ denote the representative keys of block $B$, and $\mathbf{q}_t$ be the decoding query at step $t$. Block retrieval under a budget $k$ is written as
\begin{equation}
\mathcal{S}_t(k) = \arg\max_{\substack{\mathcal{S}\subseteq\mathcal{B}\\|\mathcal{S}|=k}}
\sum_{B\in\mathcal{S}} \mathrm{sim}\!\left[\mathbf{q}_t, \mathbf{r}(B)\right],
\end{equation}
where $\mathrm{sim}[\cdot,\cdot]$ denotes the dot-product similarity between the decoding query and the representative keys of each block, and the expansion condition is formally defined as $\max(p_{t,\mathrm{hal}}, p_{t,\mathrm{unk}}) > p_{t,\mathrm{cor}}$. Algorithm~\ref{alg:refined_greedy} presents the detailed pseudocode for our UT-ACA framework.

%% file: chap/4-exp.tex
\section{Experiments}
We evaluate UT-ACA from three perspectives, including uncertainty detection, long-context generation, and cross-domain generalization. We first describe the synthetic biography dataset and evaluation metrics, then compare the learned uncertainty detector with logit-based heuristic baselines. Finally, we integrate the detector into the adaptive context allocation framework and evaluate its efficiency-quality trade-off against fixed-budget context-management methods.

\subsection{Dataset and Metrics}

\begin{figure}[t]
  \centering
  \includegraphics[width=\columnwidth]{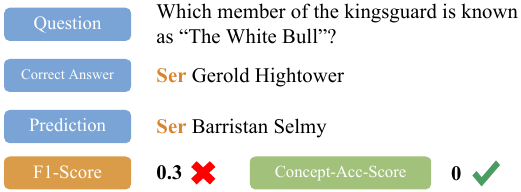}
  \caption{Example comparison between F1 score and Conceptual Accuracy Score (CAS). F1 score gives partial credit for surface overlap, while CAS assigns zero because the prediction is semantically incorrect.}
  \label{fig:concept-acc-score}
\end{figure}

\paragraph{Dataset.} We construct a synthetic biography summarization dataset using a summary-first pipeline, where each instance contains a randomly generated attribute value for a fictitious person and is expanded from a ground-truth summary into a longer biography. Person names are unique across splits to prevent data leakage. The dataset includes $10{,}000$ training samples, a validation set with six unseen attributes and $20$ summaries each, and $100$ Wikipedia-extended long-context test biographies of 171k--400k tokens. For token-level supervision, we constrain the context window during decoding, record output logits and hidden-state embeddings, and label each generated token as correct, unknown, or hallucinated using GPT-OSS-120B~\cite{openai2025gptoss} through a constrained comparison against the explicit reference fields, followed by manual verification. Further details on dataset construction, uncertainty-label generation, and representative examples are provided in Appendix~\ref{subsec:data_construction}, Appendix~\ref{app:label-construction}, and Appendix~\ref{subsec:data_examples}, respectively.

\paragraph{Metrics.} For the uncertainty detection, we report mean accuracy (\textbf{mAcc}) and per-class recall for uncertain and certain samples (\textbf{Recall}$_N$ and \textbf{Recall}$_P$). For generation quality, GPT-OSS-120B scores conceptual alignment between model outputs and ground truth, reported as Conceptual Accuracy Score (\textbf{mAcc}$_{\text{conc}}$). An illustrative comparison is shown in \figurename~\ref{fig:concept-acc-score}. Efficiency is quantified by mean context token consumption (\textbf{mTokens}) and average per-token decoding latency (\textbf{mTime}$_{\text{tok}}$).

\subsection{Uncertainty Detection Experiments}
We train and evaluate token-level uncertainty detectors on logits and hidden-state embeddings extracted from Llama-3.1-8B-Instruct~\cite{llama3herd2024}, with token labels assigned by the GPT-OSS-120B reader described above. As summarized in \tablename~\ref{tab:uncertainty-detection-comparison}, we compare UT-ACA against four uncertainty estimation baselines. \textsc{Top-10 Std} and \textsc{Logit Margin} are unsupervised logit-only heuristics, using the standard deviation of the top-10 logits and the top-1 versus top-2 margin, respectively. 
We also include two supervised evidential baselines, both formulated as binary uncertainty classifiers by merging \textit{Unknown} and \textit{Hallucination} into a single uncertain class. \textsc{Evidential Mass} uses the Dirichlet uncertainty mass, whereas \textsc{Evidential} uses the Dirichlet mean probability assigned to the uncertain class.

\begin{table}[H]
  \centering
  \fontsize{8.3pt}{10pt}\selectfont
  \setlength{\tabcolsep}{5pt}
    \begin{tabular}{lccc}
      \toprule
      \textbf{Methods} & \textbf{mAcc $({\%})$} & \textbf{Recall$_{N}$ $({\%})$} & \textbf{Recall$_{P}$ $({\%})$}\\
      \midrule
      \textsc{Top-10 Std}       & 57.43       & 35.66      & 73.19            \\
      Logit Margin              & 55.54       & 48.09      & 60.93            \\
      \midrule
      Evidential Mass           & 58.29       & 55.87      & 60.04            \\
      Evidential                & 70.85       & 74.45      & 68.25            \\
      \midrule
      \rowcolor{blue!4}
      UT-ACA                    & \textbf{83.43} & \textbf{89.71} & \textbf{81.27} \\
      \bottomrule
    \end{tabular}
  \caption{Token-level uncertainty detection performance. \textit{Unknown} and \textit{Hallucination} are grouped as uncertain, while \textit{Correct} is treated as certain. mAcc denotes classification accuracy; Recall$_N$ and Recall$_P$ denote recall on uncertain and certain tokens, respectively.}
  \label{tab:uncertainty-detection-comparison}
\end{table}

The logit-only heuristics provide weak and imbalanced token-level discrimination: \textsc{Top-10 Std} obtains higher certain-token recall but misses most uncertain tokens, while \textsc{Logit Margin} improves uncertain-token recall but remains close to chance-level mAcc. The evidential baselines benefit from supervised hidden-state features, with \textsc{Evidential} improving mAcc to $70.85\%$ and uncertain-token recall to $74.45\%$. However, UT-ACA remains substantially stronger than all baselines, reaching $83.43\%$ mAcc and $89.71\%$ recall on uncertain tokens. This gap shows that UT-ACA learns a more reliable token-level signal for detecting when the compact context window is insufficient, while preserving high recall on certain tokens.

\subsection{Generation Experiments}

We evaluate the generation performance of UT-ACA with context management baselines, and we formalize the context window update policies for UT-ACA as follows: \emph{Update: Set.n} resets the context block budget to $n$ for the next token, whereas \emph{Update: Sub.n} decreases the current budget by $n$, with a minimum budget of $1$.

\begin{table}[H]
  \centering
  \fontsize{8.3pt}{10pt}\selectfont
    \begin{tabular}{lccc}
      \toprule
      \textbf{Methods}& \textbf{Settings} & \textbf{mTokens $\downarrow$} & \textbf{mAcc$_{\text{conc}}({\%})$} $\uparrow$ \\
      \midrule
      \multirow{2}{*}{InfLLM} & $K=2$  & 32 & 91.37 \\
       & $K=3$  & 48 & \textbf{99.83} \\
      \midrule
      \rowcolor{blue!4}
      & Update:~Set.1 & 25 & 96.59 \\
      \rowcolor{blue!4}
      \multirow{-2}{*}{UT-ACA} & Update:~Sub.1  & \textbf{29} & 99.08 \\
      \bottomrule
    \end{tabular}
  \caption{Generation performance on the validation set. UT-ACA is compared with fixed-budget InfLLM in terms of average context tokens (mTokens) and conceptual accuracy (mAcc$_{\text{conc}}$), showing that adaptive allocation preserves high accuracy with fewer tokens.}
  \label{tab:val-context-acc-comparison}
\end{table}

\begin{table*}[b]
  \centering
  \fontsize{8.3pt}{10pt}\selectfont
  \setlength{\tabcolsep}{8.5pt}
  \begin{tabular}{l l c c c c} 
      \toprule
      \textbf{Models} & \textbf{Methods} & \textbf{Settings} & \textbf{mTokens} $\downarrow$ & \textbf{mTime}$_{\text{tok}}$ (s) $\downarrow$ & \textbf{mAcc}$_{\text{conc}}$ (\%) $\uparrow$ \\
      \midrule
      \multirow{23}{*}{Llama-3.1-8B-it} 
      & $^\dagger$RetrievalAttn~\cite{liu2024retrievalattention} & $K_{\text{tok}}=2048$ & 2k & 0.021 & 15.25 \\
      \cmidrule{2-6} 
      & $^\dagger$LLMLingua-2~\cite{pan-etal-2024-llmlingua} & $K_{\text{com}}=25\%$ & 32k & - & 18.08 \\
      \cmidrule{2-6}
      & $^\dagger$SnapKV~\cite{li2024snapkv} & $K_{\text{tok}}=1024$ & 1k & 0.210 & 23.43 \\
      \cmidrule{2-6}
      & TokenSelect~\cite{wu-etal-2025-tokenselect} & $K_{\text{tok}}=1024$ & 1k & - & 39.43 \\
      \cmidrule{2-6}
      & OmniKV~\cite{hao2025omnikv} & $K_{\text{tok}}=6.7\%$ & 8k & - & 40.47 \\
      \cmidrule{2-6}

      & $^\dagger$InfLLM-V2~\cite{zhao2025infllmv2densesparseswitchableattention} & $K_{\text{tok}}=1024$ & 1k & - & 40.29 \\
      \cmidrule{2-6}

      & Ada-KV~\cite{feng2025adakv} & $K_{\text{tok}}=2048$ & 2k & - & 48.89 \\
      \cmidrule{2-6}

      & FastKV~\cite{jo-etal-2026-fastkv} & $K_{\text{tok}}=4096$ & 4k & 0.212 & 51.40 \\
      \cmidrule{2-6}
      & FreqKV~\cite{kai2026freqkv} & $K_{\text{tok}}=2050$ & 2k & 0.253 & 56.94 \\
      \cmidrule{2-6}
      & H$_{2}$O~\cite{zhang2023h2o} & $K_{\text{tok}}=1024$ & 1k & 0.123 & 65.81 \\
      \cmidrule{2-6}
      
      & \multirow{3}{*}{InfLLM~\cite{xiao2024infllm}} & $K=8$  & 128 & 0.064 & 25.81 \\
      & & $K=16$ & 256 & 0.072 & 51.23 \\
      & & $K=32$ & 512 & \textbf{0.080} & 72.26 \\
      \cmidrule{2-6}
      
      \rowcolor{blue!4}
      \cellcolor{white}  
      & & $K_{\text{max}}=32$ & 123 & 0.069 & 45.23 \\
      \rowcolor{blue!4}
      \cellcolor{white} 
      & & $K_{\text{max}}=48$ & 218 & 0.075 & 62.17 \\
      \rowcolor{blue!4}
      \cellcolor{white}  
      & & $K_{\text{max}}=64$ & 344 & 0.082 & 70.48 \\
      \rowcolor{blue!4}
      \cellcolor{white} 
      & \multirow{-4}{*}{UT-ACA~(Update:~Sub.16)} & $K_{\text{max}}=96$ & \textbf{498} & 0.097 & \textbf{74.61} \\
      \midrule

      \multirow{6}{*}{Qwen2-7B-it} 
      & \multirow{3}{*}{InfLLM~\cite{xiao2024infllm}} & $K=8$  & 128 & 0.061 & 37.77 \\
      & & $K=16$ & 256 & 0.064 & 44.39 \\
      & & $K=32$ & 512 & \textbf{0.077} & 54.72 \\
      \cmidrule{2-6}
      
      \rowcolor{blue!4}
      \cellcolor{white} 
      & & $K_{\text{max}}=16$ & 119 & 0.059 & 37.08 \\
      \rowcolor{blue!4}
      \cellcolor{white} 
      & & $K_{\text{max}}=32$ & 302 & 0.066 & 48.33 \\
      \rowcolor{blue!4}
      \cellcolor{white} 
      & \multirow{-3}{*}{UT-ACA~(Update:~Sub.8)} & $K_{\text{max}}=48$ & \textbf{432} & 0.079 & \textbf{56.64} \\
      \bottomrule
    \end{tabular}
  \caption{Context-management performance on the long-context test set, where Update:~Sub.$\emph{N}$ denotes the policy of decreasing the block budget by $\emph{N}$ after each accepted token, with a minimum of 1. The table compares token selection, compression, fixed-budget retrieval, and UT-ACA across two backbones using context usage (mTokens), per-token latency (mTime$_{\text{tok}}$), and conceptual accuracy (mAcc$_{\text{conc}}$). Methods marked with $^\dagger$ exclude samples with out-of-memory errors or incoherent outputs.}
  \label{tab:test-set-exp}
\end{table*}

\paragraph{Validation set.} We compare two post-expansion update rules, \emph{Update: Set.1} and \emph{Update: Sub.1}. All methods use a fixed block size of $16$ tokens and a maximum block budget of $K_{\max}=3$. As shown in \tablename~\ref{tab:val-context-acc-comparison}, UT-ACA achieves $99.08\%$ \textbf{mAcc}$_{\text{conc}}$ with only $29$ context tokens on average, approaching InfLLM's peak performance ($99.83\%$ at $48$ tokens) while reducing context usage by approximately $40\%$. These results indicate that uncertainty-triggered expansion selectively allocates additional context without unnecessarily increasing the average context budget.

\paragraph{Long-context test set.} We further evaluate UT-ACA on a long-context test set consisting of biographies with 171k-400k tokens, with an average length of 252k tokens. These inputs exceed the nominal context lengths of both Llama-3.1-8B-Instruct and Qwen2-7B-Instruct~\cite{yang2024qwen2}, providing a challenging setting for context-management methods. For each backbone, we train a separate uncertainty detector using the same procedure as above. \tablename~\ref{tab:test-set-exp} reports context usage, per-token decoding latency, and conceptual accuracy across compression-, token-selection-, and block-retrieval-based baselines. For token-selection methods, $K_{\text{tok}}$ denotes the number of tokens selected per decoding step, while $K_{\text{com}}$ denotes the compression ratio for LLMLingua-2. InfLLM-V2 follows its original block size of $64$ with $K=16$ retrieved blocks. For a direct comparison between InfLLM and UT-ACA, both methods use a block size of $16$, where $K$ denotes the fixed block budget of InfLLM and $K_{\max}$ denotes the maximum block budget available to UT-ACA.

As shown in \tablename~\ref{tab:test-set-exp}, UT-ACA consistently achieves a favorable efficiency-quality trade-off across both backbones. On Llama-3.1-8B-Instruct, UT-ACA with $K_{\max}=96$ attains $74.61\%$ mAcc$_{\text{conc}}$ at $498$ tokens, outperforming the best InfLLM ($72.26\%$ at $512$ tokens); at $K_{\max}=64$, it reaches $70.48\%$ with only $344$ tokens, a $33\%$ reduction at a $1.78\%$ accuracy cost. On Qwen2-7B-Instruct, UT-ACA ($K_{\max}=48$) similarly exceeds InfLLM's peak ($56.64\%$ vs.\ $54.72\%$, $432$ vs.\ $512$ tokens), confirming the consistent efficiency-quality advantage of adaptive allocation. Given the test-set size of 100 biographies, we treat these differences as observed gains rather than claims of statistical significance; the primary evidence is the consistent quality--efficiency trade-off across operating points and backbones.

\subsection{Quality--Latency--Memory Trade-off}
\label{sec:quality-latency-memory}

We further evaluate the quality--latency--memory trade-off on the long-context test set using Llama-3.1-8B-Instruct. We compare fixed-budget InfLLM with $K\in\{8,16,32\}$ against UT-ACA with $K_{\max}\in\{32,48,64,96\}$. All operating points use the same backbone, GPU, prompts, and test set. End-to-end (E2E) latency combines long-context prefilling, decoding, uncertainty detection, rollback, and regeneration costs, while peak GPU memory is measured over the complete inference process.

\begin{table}[t]
\centering
\fontsize{8.3pt}{10pt}\selectfont
\setlength{\tabcolsep}{2.5pt}
\begin{tabular}{lcccc}
\toprule
\textbf{Method} &
\textbf{mAcc$_{\text{conc}}({\%})$} $\uparrow$ &
\textbf{mTok.} $\downarrow$ &
\textbf{E2E T. (s)} $\downarrow$ &
\textbf{GPU (GiB)} \\
\midrule
\multirow{3}{*}{InfLLM}
& 25.81 & 128 & 95.80 & 26.42 \\
& 51.23 & 256 & 97.46 & 26.42 \\
& 72.26 & 512 & 99.56 & 26.44 \\
\midrule
\multirow{4}{*}{UT-ACA}
& 45.23 & 123 & 98.42  & 26.44 \\
& 62.17 & 218 & 99.56  & 26.47 \\
& 70.48 & 344 & 100.70 & 26.51 \\
& \textbf{74.61} & 498 & 102.98 & 26.56 \\
\bottomrule
\end{tabular}
\caption{Quality--latency--memory trade-off on the long-context test set. InfLLM rows correspond to $K=\{8,16,32\}$ and UT-ACA rows to $K_{\max}=\{32,48,64,96\}$, ordered from top to bottom. mTok., E2E T., and GPU denote selected context tokens, end-to-end latency, and peak GPU memory, respectively.}
\label{tab:quality-latency-memory}
\end{table}

As shown in Table~\ref{tab:quality-latency-memory}, UT-ACA provides substantially stronger quality--context trade-offs with modest E2E latency overhead. At comparable context usage, $K_{\max}=32$ achieves 45.23\% mAcc$_{\mathrm{conc}}$ with 123 tokens, compared with 25.81\% for InfLLM at $K=8$ with 128 tokens. At $K_{\max}=48$, UT-ACA improves accuracy from 51.23\% to 62.17\% relative to InfLLM at $K=16$, while reducing attended context by 14.8\% (218 vs.\ 256 tokens) with only 2.10\,s additional latency. Similarly, $K_{\max}=64$ remains within 1.78 points of InfLLM at $K=32$ while reducing context usage by 32.8\% (344 vs.\ 512 tokens), with only 1.14\,s additional latency. The largest setting, $K_{\max}=96$, achieves the highest accuracy of 74.61\%, exceeding InfLLM at $K=32$ by 2.35 points while still using slightly fewer selected tokens. Peak GPU memory remains nearly unchanged (26.42--26.56\,GiB), as UT-ACA primarily adapts attended evidence rather than model-residency memory. Overall, these results demonstrate that uncertainty-triggered context allocation improves the quality--context trade-off with bounded system overhead, even on our relatively short-output biography summarization task; longer-generation settings may provide further benefits by amortizing context-management costs and correcting uncertain tokens before errors propagate across subsequent decoding steps.

\subsection{Ablation Study on Test Set}
We conduct an ablation study on the test set using Llama-3.1-8B-Instruct with $K_{\max}=64$ under the \emph{Update: Sub.16} configuration. The contribution of each component in the uncertainty detector is evaluated through systematic ablation by removing and combining the three components.

\begin{table}[h]
  \centering
  \fontsize{8.3pt}{10pt}\selectfont
    \begin{tabular}{ccccc}
      \toprule
      \textbf{LogM} & \textbf{SE} & \textbf{LSTM} & \textbf{mAcc $(\%)$} & \textbf{mAcc$_{\text{conc}}(\%)$}\\
      \midrule
      \cmark      & \xmark       & \xmark       & 26.82   &     -  \\   
      \xmark       & \cmark      & \xmark        & 78.95   &  69.58  \\      
      \cmark      & \xmark       & \cmark      & 53.71   &  -  \\      
      \xmark       & \cmark      & \cmark       & 80.06   &  68.58  \\      
      \cmark      & \cmark      & \xmark       & 80.19   &  69.42  \\      
      \cmark      & \cmark      & \cmark      & \textbf{83.43}  & \textbf{70.48}       \\
      \bottomrule
    \end{tabular}
  \caption{Component ablation of the uncertainty detector. The table evaluates the contribution of the Logit Margin branch (LogM), the Semantic Embedding branch (SE), and the LSTM module by removing them in different combinations. mAcc measures detector accuracy, and mAcc$_{\text{conc}}$ measures downstream conceptual generation accuracy.}
  \label{tab:ablation_study}
\end{table}

The results are reported in \tablename~\ref{tab:ablation_study}. Removing the Semantic Embedding branch results in substantially degraded uncertainty detection accuracy, which is insufficient to support reliable context reallocation. Consequently, generation performance under these settings is not reported. Overall, the full model that incorporates all components achieves the strongest uncertainty detection performance and consistently yields the highest generation quality across the evaluated configurations. These results suggest that semantic representations and temporal aggregation provide complementary signals that improve the stability of uncertainty estimation and enhance the reliability of the reallocation trigger mechanism.

\begin{table*}[b]
\centering
\fontsize{8.3pt}{10pt}\selectfont
\setlength{\tabcolsep}{3pt}
\begin{tabular}{l cc cc cc cc cc cc}
\toprule
\multirow{3}{*}{\textbf{Method}}
  & \multicolumn{4}{c}{\textbf{$\infty$Bench}}
  & \multicolumn{4}{c}{\textbf{LongBench}}
  & \multicolumn{4}{c}{\textbf{RULER}} \\
\cmidrule(lr){2-5}\cmidrule(lr){6-9}\cmidrule(lr){10-13}
  & \multicolumn{2}{c}{\textit{Sum Task}} & \multicolumn{2}{c}{\textit{QA Task}}
  & \multicolumn{2}{c}{\textit{narrativeqa}} & \multicolumn{2}{c}{\textit{qmsum}}
  & \multicolumn{2}{c}{\textit{QA-16k}} & \multicolumn{2}{c}{\textit{QA-8k}} \\
\cmidrule(lr){2-3}\cmidrule(lr){4-5}\cmidrule(lr){6-7}\cmidrule(lr){8-9}\cmidrule(lr){10-11}\cmidrule(lr){12-13}
  & mTok$\downarrow$ & Score$\uparrow$
  & mTok$\downarrow$ & Score$\uparrow$
  & mTok$\downarrow$ & Score$\uparrow$
  & mTok$\downarrow$ & Score$\uparrow$
  & mTok$\downarrow$ & Score$\uparrow$
  & mTok$\downarrow$ & Score$\uparrow$ \\

\midrule
& 1024 & 11.25 & 1024 & 1.14 & 64  & 5.35 & 64  & 13.84 & 128 & 12.00 & 128 & 14.30 \\
& 2048 & 11.52 & 2048 & 0.93 & 128 & 4.68 & 128 & 14.25 & 256 & 11.00 & 256 & 16.60 \\
\multirow{-3}{*}{H$_{2}$O}
& 4096 & 10.59 & 4096 & 1.27 & 256 & 4.44 & 256 & 14.81 & 512 & 10.70 & 512 & 21.00 \\

\midrule
& 1024 & 27.24 & 1024 & 16.27 & 64  & 14.95 & 64  & 18.40 & 128 & 14.05 & 128 & 14.90 \\
& 2048 & 27.87 & 2048 & 18.38 & 128 & 15.75 & 128 & 18.62 & 256 & 18.75 & 256 & 24.30 \\
\multirow{-3}{*}{InfLLM}
& 4096 & \textbf{28.58} & 4096 & \textbf{22.09} & 256 & 15.57 & 256 & 19.13 & 512 & \textbf{21.70} & 512 & 26.75 \\

\midrule
\rowcolor{blue!4}
& 609  & 27.46 & 896  & 16.58 & 39  & 15.01 & 24  & 17.65 & 48  & 10.09 & 51  & 12.60 \\
\rowcolor{blue!4}
& 1610 & 27.64 & 1664 & 18.72 & 59  & 16.11 & 47  & 18.66 & 105 & 14.65 & 108 & 14.40 \\
\rowcolor{blue!4}
\multirow{-3}{*}{\textbf{UT-ACA}}
& \textbf{3508} & 28.43 & \textbf{3712} & 21.32 & \textbf{109} & \textbf{16.92} & \textbf{102} & \textbf{19.18} & \textbf{385} & 21.65 & \textbf{352} & \textbf{26.85} \\
\bottomrule
\end{tabular}
\caption{Results on $\infty$Bench, LongBench, and RULER using Llama-3.1-8B-Instruct. The table compares H$_2$O, InfLLM, and UT-ACA under the official evaluation protocols of the corresponding benchmarks, without retraining the UT-ACA uncertainty detector. mTok denotes the average number of selected tokens, Score follows the corresponding benchmark metric, and bold values indicate the best result in each column.}
\label{tab:openset_combined}
\end{table*}

\subsection{Out-of-Domain Generalization}
As presented in Table~\ref{tab:openset_combined}, we further evaluate UT-ACA on three out-of-domain long-context benchmarks, including $\infty$Bench~\cite{zhang2024infinitebench}, LongBench~\cite{bai2024longbench}, and RULER~\cite{hsieh2024ruler}, to examine whether the learned uncertainty detector generalizes beyond the synthetic biography domain. The detector is directly reused from the version trained on our synthetic dataset, without any benchmark-specific retraining. We follow the official evaluation protocol of each benchmark and compare InfLLM with fixed context budgets $K$ against UT-ACA with corresponding maximum budgets $K_{\max}$. For $\infty$Bench, we use a block size of $128$ with $K,K_{\max}\in\{8,16,32\}$, and report the benchmark-provided scores for the summarization and QA tasks. For LongBench, we use a block size of $16$ with $K,K_{\max}\in\{4,8,16\}$, evaluating \texttt{narrativeqa} with normalized token-level F1 and \texttt{qmsum} with ROUGE-L $\times 100$. For RULER, we evaluate QA-16k and QA-8k using a block size of $16$ with $K,K_{\max}\in\{8,16,32\}$, and report the official string-match accuracy, which measures whether any normalized gold answer appears in the model prediction.

Across the evaluated operating points, the frozen uncertainty detector yields competitive quality--token Pareto efficiency despite the substantial distribution shift across benchmarks. In particular, UT-ACA consistently operates with fewer selected tokens than the corresponding fixed-budget InfLLM setting, while the associated quality loss is generally small and is often fully recovered as the maximum budget increases. On $\infty$Bench, UT-ACA remains close to InfLLM on both summarization and QA while using fewer selected tokens at every budget; at the largest budget, it achieves scores of $28.43$ and $21.32$ with $14.4\%$ and $9.4\%$ fewer tokens, respectively. The efficiency gains are more pronounced on LongBench: UT-ACA improves narrativeqa from $15.57$ to $16.92$ using $109$ rather than $256$ tokens, and obtains $19.18$ on qmsum using only $102$ tokens, compared with InfLLM's $19.13$ at $256$ tokens, corresponding to reductions of $57.4\%$ and $60.2\%$. On RULER, aggressive pruning causes larger degradation at smaller budgets, but UT-ACA recovers comparable performance at the largest budget while retaining substantially less context, achieving $21.65$ versus $21.70$ on QA-16k with $385$ versus $512$ tokens and $26.85$ versus $26.75$ on QA-8k with $352$ versus $512$ tokens.

Overall, without any benchmark-specific adaptation, UT-ACA provides competitive or non-dominated quality--efficiency operating points across all three benchmarks. These results suggest that the uncertainty detector transfers beyond its synthetic training domain and can support adaptive context allocation under substantially different task distributions, although the optimal degree of context reduction remains task dependent.

%% file: chap/5-rela.tex
\section{Related Work}
Long-context inference has been studied from two complementary directions. Rather than increasing the model's intrinsic context capacity, these methods seek to use an existing long context more efficiently by identifying, retaining, or compressing the information most relevant to generation. One line applies training-based methods~\cite{chen2025ladm, hu-etal-2025-longrecipe, gao-etal-2025-train, tian-etal-2025-untie} or improved positional encodings~\cite{ding2024longrope, jin2024llm, wang-etal-2024-resonance} to extend length generalization. Another line focuses on context management to control the effective attention scope, reduce memory and computation, and suppress irrelevant context~\cite{Liskavets_Ushakov_Roy_Klibanov_Etemad_Luke_2025, liu2024retrievalattention, Hooper2024squeezed, fu2024lazyllm}. Since UT-ACA operates at inference time by regulating the usable context, we focus on the latter.

\paragraph{Context selection.} These methods retain only the most relevant tokens or blocks to mitigate attention dilution~\cite{zhang2025PQCache, xiao2024streamingllm}. Training-free approaches such as InfLLM~\cite{xiao2024infllm} and TokenSelect~\cite{wu-etal-2025-tokenselect} perform block- or token-level retrieval over the KV cache, while structured sparse attention methods such as XAttention~\cite{xu2025xattention} and OmniKV~\cite{hao2025omnikv} design scoring-guided selection to balance efficiency and accuracy. InfLLM-V2~\cite{zhao2025infllmv2densesparseswitchableattention} and FastKV~\cite{jo-etal-2026-fastkv} further push efficiency via trainable sparse attention and accelerated eviction strategies, respectively. 
Although several methods dynamically update which tokens participate in attention during generation, their retention or sparsity budgets are typically prescribed independently of the semantic reliability of the generated token.


\paragraph{Context compression.} Rather than selecting existing entries, compression methods shrink the input or KV representations~\cite{Zhao_Wu_Xu_2025, liao-etal-2025-e2llm, wang2025sagekv, li2023selective}. Prompt-level approaches such as LLMLingua~\cite{jiang-etal-2023-llmlingua} and LongLLMLingua~\cite{jiang-etal-2024-longllmlingua} reduce token count using information-theoretic criteria. At the KV-cache level, SnapKV~\cite{li2024snapkv}, SCOPE~\cite{wu-etal-2025-scope}, and H$_2$O~\cite{zhang2023h2o} prune or evict entries based on attention patterns, while FocusLLM~\cite{li-etal-2025-focusllm} condenses long inputs into compact representations. Recent methods allow KV entries to evolve dynamically during inference, but their decisions are primarily driven by attention statistics, cache importance, or predefined compression objectives. In contrast, they do not explicitly couple context compression to token-level semantic uncertainty.


%% file: chap/appendix.tex
\section{Dataset Construction Details}

In this section, we detail the construction of the synthetic summaries and subsequent biographies. To provide insight into the data's format and diversity, we also present the attribute names and representative examples. Additionally, we include the specific prompts used to generate naturalistic biographies.

\subsection{Synthetic Dataset Construction Details}
\label{subsec:data_construction}

\begin{table*}[b]
\centering
\small
\setlength{\tabcolsep}{4pt} 
\renewcommand{\arraystretch}{1.25}
  \begin{tabular}{|p{0.16\textwidth}|p{0.27\textwidth}|p{0.5\textwidth}|}
    \hline
    \textbf{Dataset} & \textbf{Attribute Names} & \textbf{Example Data} \\
    \hline
    \textbf{Training Set} &
    \begin{minipage}[t]{\linewidth}
        \setlength{\parskip}{0pt}
        \setlength{\parindent}{0pt}
        \rowdot{Marriage Date}
        \rowdot{Job Title}
        \rowdot{Current City}
        \rowdot{Email Address}
        \rowdot{Phone Number}
        \rowdot{Favorite Color}
        \rowdot{User Agent}
        \rowdot{Credit Card Provider}
        \rowdot{Currency Used}
        \rowdot{Catch Phrase}
        \rowdot{Street Address}
        \rowdot{Vehicle License Plate}
        \rowdot{Favorite File Extension}
        \rowdot{Domain Name}
        \rowdot{Cryptocurrency}
        \rowdot{Timezone}
        \rowdot{Isbn Code}
        \rowdot{Lucky Number}
        \rowdot{Hobby}
        \rowdot{Marital Status}
        \rowdot{Personality Type}
      \newline
    \end{minipage}
    & 
    \begin{minipage}[t]{\linewidth}
        \setlength{\parskip}{0pt}
        \setlength{\parindent}{0pt}
        \rowdot{The marriage date of Laetitia Guyot de Paris is 2010-03-29.}
        \rowdot{The job title of Franck Jourdan is Magazine features editor.}
        \rowdot{The current city of Marcial Escalona Yuste is Augerdan.}
        \rowdot{The email address of Jeremy Wright is catherine40@example.com.}
        \rowdot{The phone number of Clarence Byrd is 07703087971.}
        \rowdot{The favorite color of Letizia Fagotto is Black.}
        \rowdot{The user agent of Inés Novoa Sarabia is Mozilla/5.0 }
        \rowdot{The credit card provider of Douglas Baker is Maestro.}
        \rowdot{The currency used of Susann Jacobi Jäckel is Bahraini dinar.}
        \rowdot{The catch phrase of Olga Torre Rodriguez is Interfaccia estesa ...}
        \rowdot{The street address of Gabriela Jesus is Bährgasse 62.}
        \rowdot{The vehicle license plate of Erdal Mälzer is GE 3041 CB.}
        \rowdot{The favorite file extension of Ornella Tomasini is tiff.}
        \rowdot{The domain name of Théo Macedo is chretien.com.}
        \rowdot{The cryptocurrency of Anaïs Gilbert du Rousset is Vertcoin.}
        \rowdot{The timezone of Eutimio Tejada Ibañez is Europe/Paris.}
        \rowdot{The isbn code of Jeanne-Sabine Brunet is 978-0-456-54959-9.}
        \rowdot{The lucky number of Zoé Chauvet is 61.}
        \rowdot{The hobby of María Manuela Guardia Moles is Sleeping.}
        \rowdot{The marital status of Gustavo Vallés is Widowed.}
        \rowdot{The personality type of Salvi Toscanini-Ferrabosco is Extrovert.}
        \newline
    \end{minipage}
    \\
    \hline
    \textbf{Validation Set and Test Set} &
    \begin{minipage}[t]{\linewidth}
        \setlength{\parskip}{0pt}
        \setlength{\parindent}{0pt}
        \rowdot{Birth Date}
        \rowdot{Birth Place}
        \rowdot{Company}
        \rowdot{Major}
        \rowdot{University}
        \rowdot{Work Place}
        \newline
    \end{minipage}
    & 
    \begin{minipage}[t]{\linewidth}
        \setlength{\parskip}{0pt}
        \setlength{\parindent}{0pt}
        \rowdot{The birth date of Olivia Garcia is March 22, 1985.}
        \rowdot{The birth place of Daniel Lee is Seoul, South Korea.}
        \rowdot{The company of Olivia Chen is Microsoft.}
        \rowdot{The major of Ethan Clark is Computer Science.}
        \rowdot{The university of Aiden Gomez is Stanford University.}
        \rowdot{The work place of Elijah Robinson is Singapore, Singapore.}
        \newline
    \end{minipage}
    \\
    \hline
  \end{tabular}
  \caption{Attribute sets and example facts in the synthetic dataset. The table lists representative attribute names and example facts for the training, validation, and test splits. Training uses diverse single-attribute facts, while validation and test focus on six held-out biography attributes.}
  \label{tab:example_attributes}
\end{table*}

\begin{table*}[t]
\centering
\small
\setlength{\tabcolsep}{3pt}
\renewcommand{\arraystretch}{1.25}
\begin{tabular}{lllll}
\toprule
\multicolumn{5}{c}{\textbf{Wikipedia Source Articles}} \\
\midrule
Architecture & Mathematics & Pacific Ocean & Thermodynamics & Blockchain \\
Art & Number & Sahara & Virus & Cryptography \\
Baroque & Sport & South America & Communism & Engineering \\
Cinema & Vaccine & Volcano & Culture & Internal combustion engine \\
Classical music & French Revolution & Age of Enlightenment & Democracy & Internet \\
Impressionism & Maya civilization & American Civil War & Epistemology & Nanotechnology \\
Jazz & Roman Empire & Ancient Egypt & Human rights & Robotics \\
Literature & Africa & Ancient Greece & Law & Semiconductor \\
Modernism & Amazon River & British Empire & Logic & Software \\
Mona Lisa & Amazon rainforest & Bacteria & Metaphysics & Steam engine \\
Painting & Antarctica & Biology & Philosophy & Technology \\
Photography & Atlantic Ocean & Black hole & Psychology & Telecommunication \\
Poetry & Australia (continent) & Cell (biology) & Republic & Agriculture \\
Rock music & Earth & Climate change & Socialism & Calendar \\
Sculpture & Europe & DNA & Ottoman Empire & Food \\
The Starry Night & Great Barrier Reef & Evolution & World War II & Light \\
Theatre & Nile & General relativity & Algorithm & Medicine \\
Geometry & Ocean & Quantum mechanics & Artificial intelligence & Olympics \\
\bottomrule
\end{tabular}
\caption{Wikipedia source articles used for test-set context extension. These passages are inserted around synthetic biographies to create long distractor contexts, testing whether methods can retrieve relevant biography facts from long, heterogeneous inputs.}
\label{tab:wiki_data_name}
\end{table*}

\begin{figure}[H]
  \begin{center}
    \includegraphics[width=\columnwidth]{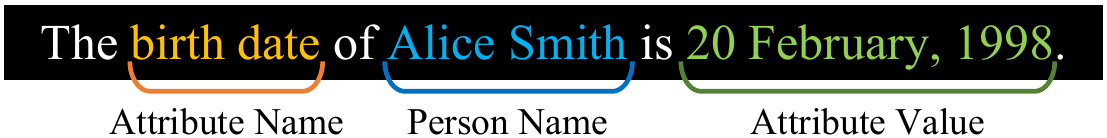}
  \end{center}
  \caption{Standardized pattern of synthetic summaries. The figure illustrates the controlled person--attribute--value structure used to define ground-truth facts. This standardized structure keeps the ground truth unambiguous while allowing the surrounding biography text to vary in wording and length.}
  \label{fig:example_syn_data}
\end{figure}

Our synthetic dataset is designed primarily for the biography summarization task. As illustrated in \figurename~\ref{fig:example_syn_data}, we employ a fixed pattern to construct the ground truth summaries. Specifically, each summary consists of a sequence with three components: the person name, the attribute name, and the attribute value. 
\textbf{It is important to note that all person names and attribute values are entirely fictitious. Therefore, there are no privacy concerns regarding real individuals.}
To generate the full biography, the summary is extended with filler content or paraphrased into a natural narrative. The dataset is partitioned into training, validation, and test sets, all constructed using synthetic summaries.

\paragraph{Training and Validation Set.}
For the training and validation data construction, each summary is associated with a single attribute, whereas the test set involves six different attributes. To ensure rigorous evaluation, the attributes used in the training set are distinct from those in the validation and test sets. \tablename~\ref{tab:example_attributes} lists all attribute names used in our synthetic dataset. Specifically, we utilize 21 attributes for training, and reserve 6 distinct attributes for validation and testing. Additionally, we provide examples for each attribute to demonstrate the diversity of the data.

\begin{figure*}[t]
  \begin{center}
    \includegraphics[width=\textwidth]{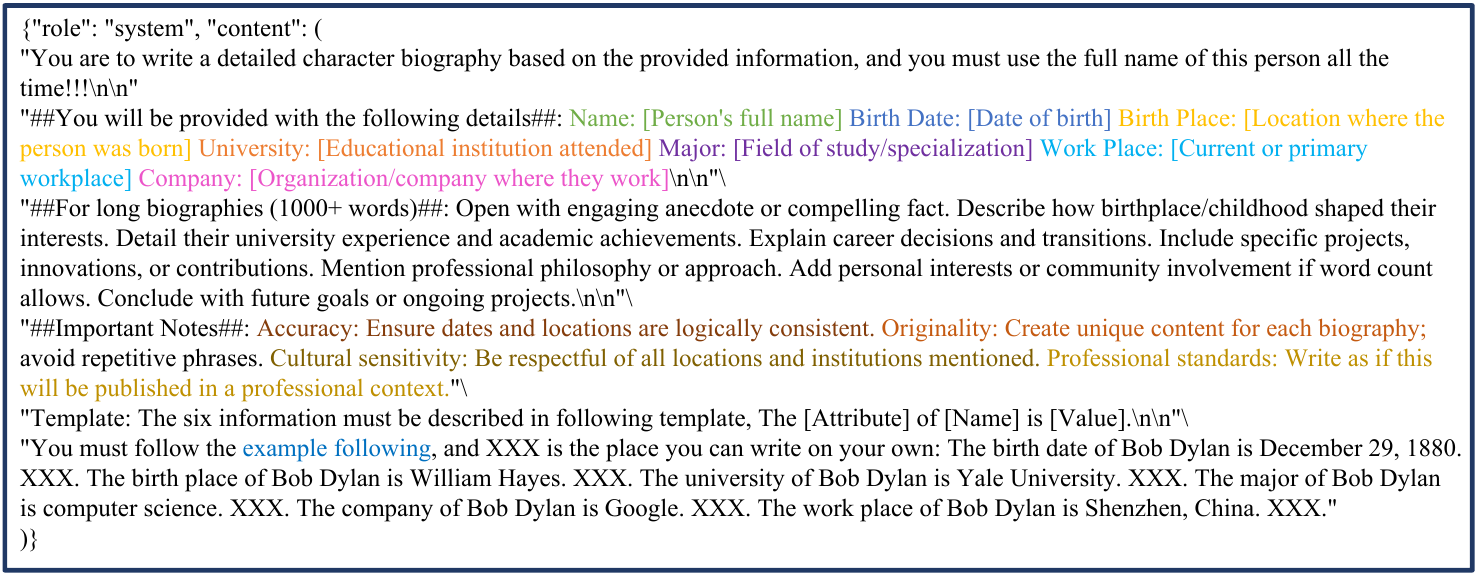}
  \end{center}
  \caption{Prompt template for generating naturalistic biographies. The figure shows the LLM instruction used to convert structured ground-truth facts into fluent biography passages. The prompt asks the LLM to preserve the provided summary facts while embedding them in natural narrative text.}
  \label{fig:llm_prompt_generate_bio}
\end{figure*}

\paragraph{Test Set.}
We employ GPT-OSS-120B to generate naturalistic biographies based on the summaries provided in the test set. It is important to note that this generation process is model-agnostic; therefore, any capable generative model can be utilized for this task. The specific prompt used to construct the biographies is shown in \figurename~\ref{fig:llm_prompt_generate_bio} and can be readily copied and applied.
After the module generate naturalistic biographies, we insert common knowledge texts from Wikipedia to further extend the length of synthetic biographies for test set. As shown in \tablename~\ref{tab:wiki_data_name}, we present the Wikipedia files used in our test set extending process.
The original link of these data can be found in \url{https://en.wikipedia.org/wiki}.

\subsection{Uncertainty Label Construction and Verification}
\label{app:label-construction}

The synthetic uncertainty labels are constructed under a controlled reference-based protocol. As described in Appendix~\ref{subsec:data_construction}, each synthetic instance is associated with an explicit ground-truth fact represented by a person name, attribute name, and attribute value. This structure provides an unambiguous reference for determining whether generated content is supported by the available evidence.

\begin{figure*}[t]
\centering
\includegraphics[width=\textwidth]{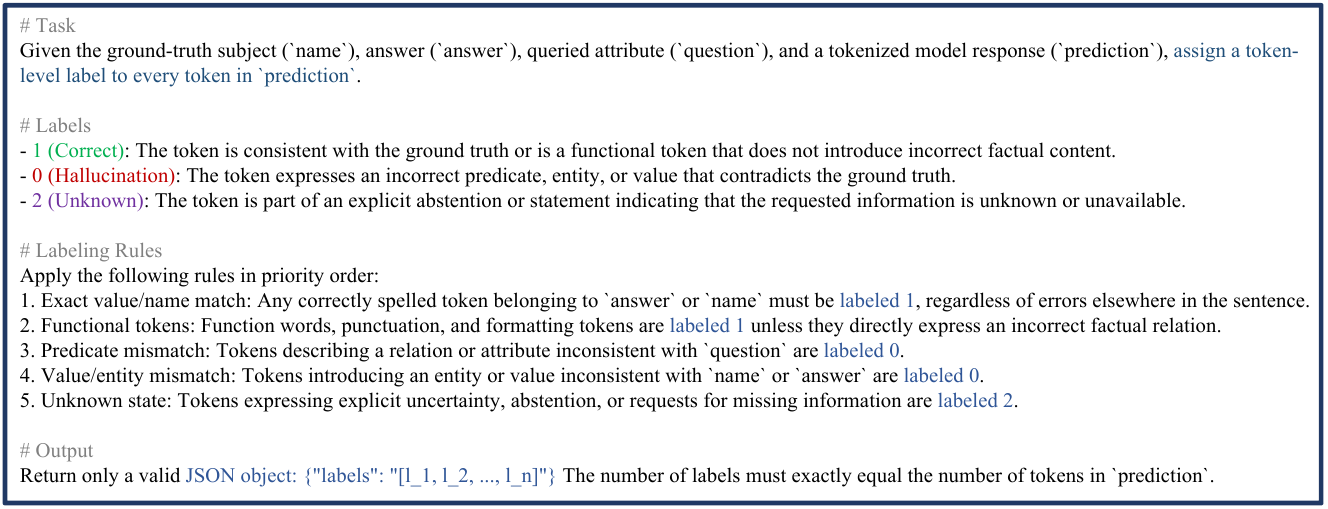}
\caption{Prompt for reference-constrained token-level uncertainty labeling. GPT-OSS-120B assigns each prediction token as \textit{Correct} (1), \textit{Hallucination} (0), or \textit{Unknown} (2) by comparing it with the explicit ground-truth subject, attribute, and answer under prioritized labeling rules.}
\label{fig:labeling-prompt}
\end{figure*}

For each generation trajectory, token-level labels are assigned according to three generation states. Tokens consistent with the reference fact are labeled \textit{Correct}; tokens that explicitly abstain from providing the requested information are labeled \textit{Unknown}; and tokens that introduce factual content unsupported by the reference are labeled \textit{Hallucination}. For template-defined tokens whose correctness is deterministic from the construction process, labels are assigned directly. For factual output tokens, GPT-OSS-120B performs a constrained comparison between the generated response and the corresponding reference fields rather than an unconstrained factuality judgment. The complete labeling prompt is shown in Figure~\ref{fig:labeling-prompt}.

The resulting labels are manually verified to identify incorrect assignments and ambiguous cases. Because the underlying facts are explicitly specified during data construction, verification primarily checks consistency between the generated tokens and the reference fields rather than requiring subjective assessment of open-ended factuality. This controlled labeling procedure reduces ambiguity in the supervision used to train the uncertainty detector.

\subsection{Data Examples from Our Dataset}
\label{subsec:data_examples}
In our experiments, the training set is utilized to train the uncertainty detector; consequently, the data structure and sequence lengths in this set are relatively simple and short. The validation set is designed to preliminarily evaluate the trained uncertainty detector and the downstream adaptive context window. While the validation data retains the same format as the training set, it comprises distinct attributes and person names. Finally, the test set is constructed to assess our method in a long-context setting. To achieve this, we insert extensive passages from Wikipedia to extend biography lengths, thereby challenging the long-context handling capabilities of the methods.

\begin{table*}[t]
\centering
\fontsize{8.3pt}{10pt}\selectfont
\renewcommand{\arraystretch}{1.25}
  \begin{tabular}{|p{0.10\textwidth}|p{0.38\textwidth}|p{0.42\textwidth}|}
    \hline
    \textbf{Dataset} & \textbf{Example Summarization} & \textbf{Example Biography} \\
    \hline

    \textbf{Training Set} &
    The \textbf{email address} of Jeremy Wright is catherine40@example.com. &
    The sky is really blue. ... The \textbf{email address} of Jeremy Wright is catherine40@example.com. ... The sky is really blue. \\
    \hline

    \textbf{Validation Set} &
    The \textbf{birth date} of Olivia Garcia is March 22, 1985. &
    The sky is really blue. ... The \textbf{birth date} of Olivia Garcia is March 22, 1985. ... The sky is really blue.\\
    \hline

    \textbf{Test Set} &
    \begin{minipage}[t]{\linewidth}
      \setlength{\parskip}{0pt}
      \setlength{\parindent}{0pt}
      \rowdot{The \textbf{birth date} of Emma Thompson is 12/29/1880.}
      \rowdot{The \textbf{birth place} of Emma Thompson is Tokyo, Japan.}
      \rowdot{The \textbf{university} of Emma Thompson is Harvard University.}
      \rowdot{The \textbf{major} of Emma Thompson is Computer Science.}
      \rowdot{The \textbf{company} of Emma Thompson is Google.}
      \rowdot{The \textbf{work place} of Emma Thompson is San Francisco Bay Area.}
      \newline
    \end{minipage}
    & ... Emma Thompson was \textbf{born on} December 29, 1880, \textbf{in the} bustling metropolis of Tokyo, Japan. ... She \textbf{applied to} Harvard University and was \textbf{accepted into} the Computer Science program. ... After careful consideration, she chose to \textbf{join} Google, a company known for its innovative culture and cutting-edge technology. ... Emma Thompson's \textbf{journey from} Tokyo \textbf{to the} San Francisco Bay Area is a testament to her unwavering determination and exceptional talent. ...
     \\
    \hline

  \end{tabular}
  \caption{Representative examples from the training, validation, and test splits. The table shows how the three splits differ in attribute complexity and context length. Training and validation examples contain one explicit fact in a short template-like passage, whereas test examples contain six biography attributes expressed in natural language and surrounded by long-context distractors.}
  \label{tab:example_data}
\end{table*}

\tablename~\ref{tab:example_data} presents examples from our synthetic dataset. While the training and validation sets involve processing only a single attribute per instance, the test set requires the simultaneous extraction of six attributes. Furthermore, since the biographies in the test set are composed in natural language, the model must identify and extract multiple attributes from natural-language context to derive the correct answers.

\section{Implementation Details}

This section first introduces training details and the architecture of our uncertainty detector. Subsequently, we describe the methodology for extracting LLM internal states used in this work. Finally, we detail the selection of network hyperparameters used in our approach.

\subsection{Training Details}

We implement UT-ACA in Python 3.11 using PyTorch 2.4. The detector is trained on a single NVIDIA H100 GPU. We conduct eight independent runs, each requiring approximately 1.5 GPU-hours. Each run performs 10,000 optimization steps with a batch size of 16.

We optimize the detector using Adam with an initial learning rate of $10^{-3}$ and no weight decay. The learning rate follows cosine annealing over the 10,000 training steps, with the minimum learning rate set to zero. We use token-level weighted cross-entropy loss and exclude padded token positions from the loss. The class weights are $[1.0, 0.05, 1.0]$ for the three-class detector, corresponding to hallucinated, correct, and unknown tokens, respectively. For the binary ablation, unknown tokens are merged into the uncertain class and the corresponding weights are $[1.0, 0.05]$.

We evaluate the detector every 100 optimization steps and retain the checkpoint with the highest validation F1 score. We do not apply early stopping: optimization continues for all 10,000 steps, while checkpoint selection is performed retrospectively using validation F1.

\subsection{Architecture of Uncertainty Detector}
\label{subsec:str_ucertaintydet}
The proposed model, termed \emph{Uncertainty Detector}, is designed to process sequential data consisting of high-dimensional semantic feature vectors and scalar confidence scores. The architecture comprises three primary modules: dual-branch feature encoding, temporal modeling, and a residual classification head.

\paragraph{Input Encoding and Fusion}
At each time step $t$, the model accepts two input streams: a feature vector $\mathbf{emb}_t \in \mathbb{R}^{D_{\text{in}}}$ (where $D_{\text{in}}=4096$ for Llama-3.1-8B) and a scalar score $m_t \in \mathbb{R}$. Before entering the semantic encoder, each feature vector is L2-normalized along its feature dimension. The scalar $m_t$ represents the logit margin, calculated by subtracting the top-2 logit from the top-1 logit of the LLM output. The vector input is processed by the \texttt{VecEncoder}, which projects the high-dimensional input to a latent space of dimension $d_{model}=64$. This is achieved via a linear transformation followed by a GELU activation and dropout
\begin{equation}
    \mathbf{h}_{vec}^{(t)} = \text{Dropout}(\text{GELU}(\mathbf{W}_e \mathbf{emb}_t + \mathbf{b}_e))
\end{equation}
where dropout rate is set to $0.5$. 
Simultaneously, the scalar score is processed by the \texttt{ScoreEncoder}, a Multi-Layer Perceptron (MLP) aimed at up-sampling the scalar to match $d_{model}$. The MLP consists of two linear layers with an intermediate dimension of 32, utilizing GELU activations and dropout ($p=0.1$).

The \texttt{ScoreEncoder} acts as the uncertainty signal branch, employing the logit margin as a lightweight indicator to gauge the generation difficulty of tokens.
Conversely, the \texttt{VecEncoder} serves as the semantic feature branch, utilizing the hidden states of the LLM to extract conceptual information for enhanced uncertainty estimation. The dual-encoder module aligns the dimensions of these two information sources, after which the features are fused via element-wise addition to form a joint representation ($\mathbf{z}_t$).

\paragraph{Temporal Modeling}
To effectively model the accumulation of uncertainty throughout the generation process, we employ a temporal modeling approach designed to capture the sequential dependencies between the current token and its predecessors. Specifically, the fused feature sequence $\mathbf{U} = \{\mathbf{z}_1, \dots, \mathbf{z}_T\}$ serves as input to a standard Long Short-Term Memory (LSTM) layer. Configured with a hidden size of $d_{model}$, the LSTM recursively processes the sequence according to LSTM dynamics.

For the training phase of the uncertainty detector, we preserve the entire sequence of hidden states as supervision signals. In contrast, during the LLM inference stage, only the final hidden state $\mathbf{h}_T$ is extracted to predict the category of the current token, as it aggregates the information of the complete sequence.

\paragraph{Residual Classification Head}
The output from the LSTM passes through a stack of $N=3$ \texttt{ResidualBlock}s. Each block implements a bottleneck structure with an expansion factor of 2. For an input $\mathbf{h}$, the block operation is defined as
\begin{equation}
\begin{aligned}
    \hat{\mathbf{h}} &= \text{Dropout}(\text{GELU}(\mathbf{W}_1 \mathbf{h} + \mathbf{b}_1)) \\
    \mathbf{h}_{out} &= \mathbf{h} + \text{Dropout}(\mathbf{W}_2 \hat{\mathbf{h}} + \mathbf{b}_2) \\
\end{aligned}
\end{equation}
Finally, a linear projection maps the output of the last residual block to three class logits, followed by a softmax function to obtain the GDM probabilities $p_{t,\mathrm{hal}}, \mathbf{p}_t=[p_{t,\mathrm{cor}},p_{t,\mathrm{unk}}]$.

\paragraph{Hyperparameters}
The specific configurations for our uncertainty detector are detailed in \tablename~\ref{tab:hyperparams}. Notably, while most hyperparameters remain consistent across different LLMs, certain settings are model-specific. For instance, the feature vector dimension is inherited directly from the evaluated LLM; thus, it is set to 3584 for Qwen2-7B. All other internal parameters of the uncertainty detector are kept uniform to ensure comparability.

\begin{table}[t]
    \centering
    \fontsize{8.3pt}{10pt}\selectfont
    \renewcommand{\arraystretch}{0.8}
    \begin{tabular}{l l c}
        \toprule
        \textbf{Component} & \textbf{Parameter} & \textbf{Value} \\
        \midrule
        \multirow{3}{*}{Input Dimensions} & Vector Dim ($D_{in}$) & 4096 \\
         & Score Dim & 1 \\
         & Hidden Model Dim ($d_{model}$) & 64 \\
        \midrule
        \multirow{2}{*}{Encoders} & Vec Dropout & 0.5 \\
         & Score MLP Inner Dim & 32 \\
        \midrule
        \multirow{3}{*}{Residual Head} & Number of Blocks & 3 \\
         & Expansion Factor & 2 \\
         & Dropout & 0.1 \\
        \midrule
        Output & Number of Classes & 3 \\
        \bottomrule
    \end{tabular}
    \caption{Hyperparameter configuration of the Llama-3.1-8B uncertainty detector. The table reports the input dimensions, encoder settings, residual-head configuration, and number of output classes. The detector projects high-dimensional LLM features into a compact 64-dimensional hidden space before classification.}
    \label{tab:hyperparams}
\end{table}

\subsection{Inner State Extraction of LLMs}

\begin{figure}[t]
  \centering
  \includegraphics[width=\columnwidth]{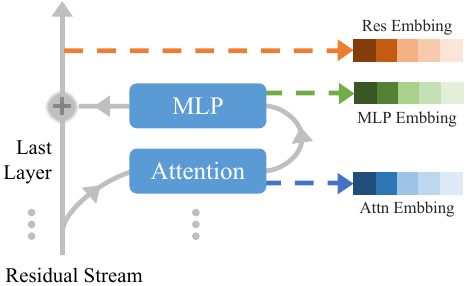}
  \caption{Extraction points for semantic embeddings in the final transformer layer. The figure marks the attention output, MLP output, and residual stream in the final layer (e.g., Layer 32 of Llama-3.1-8B). These positions are evaluated as candidate internal states for token-level semantic features.}
  \label{fig:extracted_embeddings}
\end{figure}

To train the uncertainty detector, we leverage internal embeddings from the LLM to capture token-level semantic complexity. Following established protocols, we extract features from the final layer of LLMs. As illustrated in Figure~\ref{fig:extracted_embeddings}, we consider three primary extraction points: (1) the attention mechanism output, (2) the Multi-Layer Perceptron (MLP) block output, and (3) the final residual stream output. We evaluate the efficacy of each position by training independent detectors for comparison. 

Our results indicate that the extraction site has a negligible impact on detection performance. This consistency stems from the additive nature of the residual stream, where the transformations at these sub-layer stages act as incremental refinements to the same underlying representation, making them may explain the limited sensitivity for uncertainty detection. Based on these results, we adopt the attention output (option 1) as the default extraction point in all experiments, as it is the most directly accessible in standard implementations.

\section{Evaluation Metrics Details}

In this section, we detail the implementation of the Conceptual Accuracy Score (CAS) proposed in this study. We then evaluate the reliability of CAS within the context of biography summarization by comparing it against existing evaluation methods and human annotations. Our human study provides preliminary evidence that CAS can capture conceptual correctness in this controlled biography summarization setting. Given the limited sample size, we treat this analysis as a sanity check rather than definitive validation of LLM-judge reliability.

\subsection{Details of Conceptual Accuracy Score}
\label{subsec:eval_conc_acc}

\begin{figure*}[b]
  \begin{center}
    \includegraphics[width=\textwidth]{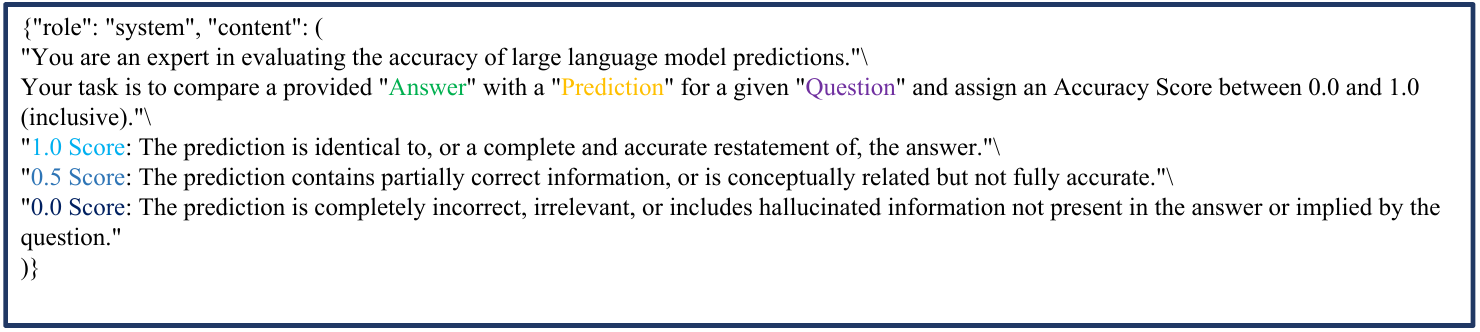}
  \end{center}
  \caption{Prompt template for evaluating Conceptual Accuracy Score (CAS). The figure shows the LLM-judge instruction used to score semantic correctness rather than surface overlap. For each subtask, the prompt provides the question, the model prediction, and the ground-truth answer, then requests a floating-point score between 0 and 1 that reflects conceptual accuracy.}
  \label{fig:llm_prompt_evaluate_acc}
\end{figure*}

In our work, we employ GPT-OSS-120B, which can theoretically be replaced by any sufficiently capable generative model, to estimate the Conceptual Accuracy Score (CAS). CAS is computed by comparing the LLM-generated output with the ground-truth summary under the specified attributes. The prompt used for this estimation is shown in \figurename~\ref{fig:llm_prompt_evaluate_acc}. The model takes as input the predictions of the evaluated LLMs, together with the corresponding questions and ground-truth answers, and outputs a floating-point score between 0 and 1, indicating the degree of accuracy of the LLM responses.

\begin{table*}[t]
    \centering
    \fontsize{8.3pt}{10pt}\selectfont
    \begin{tabular}{llccccllcccc}
    \toprule
    \textbf{ID} & \textbf{Name} & \textbf{Str} & \textbf{CAS} & \textbf{Hum} & \textbf{Cor} &
    \textbf{ID} & \textbf{Name} & \textbf{Str} & \textbf{CAS} & \textbf{Hum} & \textbf{Cor} \\
    \cmidrule(r){1-6} \cmidrule(l){7-12} 
    1  & Emma Thompson    & .000 & .000 & .000 & 0/6 & 16 & Lucas Scott      & .262 & 1.00 & 1.00 & 6/6 \\
    2  & Liam Carter      & .127 & .167 & .167 & 1/6 & 17 & Harper Coleman   & .114 & .167 & .167 & 1/6 \\
    3  & Olivia Bennett   & .230 & 1.00 & 1.00 & 6/6 & 18 & Henry Green      & .260 & .500 & .667 & 3/6 \\
    4  & Noah Wright      & .278 & .830 & .833 & 5/6 & 19 & Evelyn Perry     & .308 & 1.00 & 1.00 & 6/6 \\
    5  & Ava Sullivan     & .299 & .667 & .667 & 4/6 & 20 & Oliver Hughes    & .358 & 1.00 & 1.00 & 6/6 \\
    6  & Ethan Brooks     & .364 & .833 & .833 & 5/6 & 21 & Abigail Kelly    & .333 & .833 & .833 & 5/6 \\
    7  & Sophia Reed      & .341 & .830 & 1.00 & 6/6 & 22 & Jack Ross        & .268 & .830 & .833 & 5/6 \\
    8  & Mason Foster     & .389 & 1.00 & 1.00 & 6/6 & 23 & Emily Howard     & .355 & 1.00 & 1.00 & 6/6 \\
    9  & Isabella Parker  & .246 & .830 & .833 & 5/6 & 24 & Daniel Gray      & .311 & .830 & .833 & 5/6 \\
    10 & William Hayes    & .231 & .500 & .500 & 3/6 & 25 & Sofia Ellis      & .159 & .330 & .417 & 2/6 \\
    11 & Mia Evans        & .181 & .830 & .830 & 5/6 & 26 & Matthew Bell     & .379 & 1.00 & 1.00 & 6/6 \\
    12 & James Turner     & .385 & .833 & .830 & 5/6 & 27 & Aria Wood        & .417 & 1.00 & 1.00 & 6/6 \\
    13 & Amelia Fisher    & .227 & .833 & .833 & 5/6 & 28 & Jacob King       & .271 & .830 & .917 & 5/6 \\
    14 & Benjamin Ward    & .350 & .500 & .500 & 3/6 & 29 & Scarlett Price   & .242 & .750 & .750 & 4/6 \\
    15 & Charlotte Morgan & .323 & .830 & .830 & 5/6 & 30 & Logan Lee        & .361 & .830 & .917 & 5/6 \\
    \midrule
    \multicolumn{12}{c}{\emph{Avg (1--30):}{\hspace{2em}} Str=0.279,{\hspace{2em}} CAS=\textbf{0.746},{\hspace{2em}} Hum=\textbf{0.766},{\hspace{2em}} Cor=4.5/6} \\
    \bottomrule
    \end{tabular}
    \caption{Preliminary human comparison of Conceptual Accuracy Score on biography extraction. The table compares string matching (\textbf{Str}), conceptual accuracy score (\textbf{CAS}), and human annotation (\textbf{Hum}) on 30 test samples. \textbf{Cor} reports the number of correctly extracted attributes out of six according to human annotators.}
    \label{tab:annotation_results}
\end{table*}

\subsection{Evaluations of Conceptual Accuracy Score}

To provide a preliminary assessment of the proposed CAS, we conduct a human annotation study on 30 samples drawn from our test-set evaluation.
Specifically, human annotators are asked to evaluate $30$ samples drawn from the real evaluation workload in our test set experiments. The annotators receive the same prompt used with GPT-OSS-120B and are instructed to assign a floating-point score independently. We then compare the results obtained from three evaluation methods: string-based matching, human evaluation, and CAS estimated by GPT-OSS-120B. As each test sample requires the extraction of six attributes, we report both the average accuracy scores under the three evaluation settings and the number of correctly identified attributes, as determined by the human annotators.

\paragraph{Quantitative Evaluation.}
We randomly select $30$ samples from our test set to conduct a comparative analysis. The experimental setup is as follows: we utilize Llama-3.1-8B-Instruct as the model to be evaluated and employ our UT-ACA as the context management method for long-context inference. The specific parameters of the UT-ACA are set to a fixed block size of 16 tokens, with a maximum budget of 96 blocks for the adaptive context allocation. The update policy involves subtracting 16 blocks after the generation of tokens with certainty, maintaining a minimum window size of 1 block. As shown in Table~\ref{tab:annotation_results}, CAS obtains an average score of 0.746, compared with 0.766 from human evaluation, corresponding to an absolute difference of 0.02. 

\paragraph{Qualitative Evaluation.}
\tablename~\ref{tab:case_study} presents representative examples illustrating the behavior of CAS under surface-form variation and genuine semantic errors.
Specifically, in the evaluation samples, CAS can automatically detect formatting differences while correctly capturing the underlying semantic correctness. For example, in \textbf{Case~\#3}, the LLM predicts ``1888-06-03'' whereas the ground truth is ``June~3,~1888''. A string-matching baseline would judge this prediction as incorrect (or assign it an artificially low score) due solely to the surface-form discrepancy. In contrast, CAS correctly assigns a score of~1. In \textbf{Case \#17}, CAS accurately identifies genuine semantic errors and appropriately assigns a a low score of $0.167$.

\begin{table*}[t]
\centering
\fontsize{8.3pt}{10pt}\selectfont
\renewcommand{\arraystretch}{1.2} 
\begin{tabular}{lllc !{\qquad} lllc}
\toprule
\multicolumn{4}{c}{\textbf{Case \#3: Olivia Bennett}} & \multicolumn{4}{c}{\textbf{Case \#17: Harper Coleman}} \\
\multicolumn{4}{c}{\small (Str: 0.230, LLM: 1.000, Hum: 1.000)} & \multicolumn{4}{c}{\small (Str: 0.114, LLM: 0.167, Hum: 0.167)} \\
\cmidrule(r){1-4} \cmidrule(l){5-8}
\textbf{Attr} & \textbf{GT} & \textbf{Pred.} & \textbf{Hum} & \textbf{Attr} & \textbf{GT} & \textbf{Pred.} & \textbf{Hum} \\
\midrule
Birth Date  & June 3, 1888    & 1888-06-03              & \cmark & Birth Date  & 10 Nov, 1907      & 1976                 & \xmark \\
Birth Place & Sydney, NSW   & Sydney, NSW               & \cmark & Birth Place & Mexico City       & Madrid, Spain        & \xmark \\
University  & MIT           & Mass. Inst. Tech.         & \cmark & University  & NUS               & Complutense Madrid   & \xmark \\
Major       & Medicine      & Medicine                  & \cmark & Major       & Intl. Rel.        & Chem. Eng.           & \xmark \\
Company     & Mayo Clinic   & Mayo Clinic               & \cmark & Company     & UN                & UN                   & \cmark \\
Work Place  & Beijing, CN   & Mayo's Beijing            & \cmark & Work Place  & Seattle, WA       & Pacific NW           & \xmark \\
\bottomrule
\end{tabular}

\caption{Case study of CAS under format variation and semantic errors. Case~\#3 shows that date and entity-format differences can still be semantically correct, whereas Case~\#17 contains wrong attribute values that should receive low scores. \cmark\ and \xmark\ indicate human judgments for each attribute.}
\label{tab:case_study}
\end{table*}

Importantly, CAS is used only as a post-hoc measure of generation quality and does not participate in uncertainty detection, context allocation, or decoding. It therefore does not affect the reported context consumption or latency. Moreover, our out-of-domain experiments on $\infty$Bench,
LongBench, and RULER follow the official evaluation metrics of the corresponding benchmarks rather than CAS.

\section{Additional Results and Analysis}

\subsection{Ablation Study on Two-Way vs.\ Three-Way Difficulty Metric}

To examine the contribution of the proposed three-way difficulty metric, we conduct an ablation study in which the \textit{Unknown} and \textit{Hallucination} categories are merged into a single class, yielding a two-way difficulty metric. Under this setting, we perform eight independent training runs using the same experimental protocol. For each run, the checkpoint achieving the highest F1 score on the validation set is retained. The selected detectors are then evaluated on the test set with the following configuration: $\texttt{MaxBudget} = 64$, $\texttt{block\_size} = 16$, and $\texttt{SubPolicy} = 16$.

\begin{table*}[h]
\centering
\fontsize{8.6pt}{10.4pt}\selectfont
\begin{tabular}{lccccc}
  \toprule
  \textbf{Setting} & \textbf{mAcc (\%)} $\uparrow$ & \textbf{Recall$_\text{N}$ (\%)} $\uparrow$ & \textbf{Recall$_\text{P}$ (\%)} $\uparrow$ & \textbf{mTokens} $\downarrow$ & \textbf{mAcc$_\text{conc}$ (\%)} $\uparrow$ \\
  \midrule
  Two-Way Metric   & 82.74 \textcolor{red}{($\downarrow$0.69)} & 88.04 \textcolor{red}{($\downarrow$1.67)} & 80.60 \textcolor{red}{($\downarrow$0.67)} & 348 & 67.99 \textcolor{red}{($\downarrow$2.49)} \\
  \rowcolor{blue!4}
  Three-Way Metric & \textbf{83.43} & \textbf{89.71} & \textbf{81.27} & \textbf{344} & \textbf{70.48} \\
  \bottomrule
\end{tabular}
\caption{Ablation study of two-way versus three-way difficulty metrics. The two-way metric merges Unknown and Hallucination into one negative class, while the three-way metric keeps them distinct. mAcc denotes detector accuracy, Recall$_\text{N}$ and Recall$_\text{P}$ denote recall on negative and positive samples, mTokens denotes average context token consumption, and mAcc$_\text{conc}$ denotes conceptual generation accuracy.}
~\label{tab:ablation_metric}
\end{table*}

As shown in \tablename~\ref{tab:ablation_metric}, merging \textit{Unknown} and 
\textit{Hallucination} into a single category leads to consistent performance degradation across both detection and downstream generation metrics. At the detector level, mean accuracy (mAcc) decreases by 0.69 points (83.43\% $\to$ 82.74\%), and the recall for negative samples 
drops by 1.67 points (89.71\% $\to$ 88.04\%). More critically, the degradation propagates to the downstream generation stage: conceptual generation accuracy (mAcc$_\text{conc}$) declines substantially by 2.49 points (70.48\% $\to$ 67.99\%), while average context token consumption 
remains nearly unchanged (344 vs.\ 348 tokens), confirming that the gain is attributable to improved detection quality rather than increased token expenditure.

This degradation stems from the distinct generative regimes underlying \textit{Unknown} and \textit{Hallucination} responses. Under compact context windows, the model tends to produce \textit{Unknown} responses with disproportionately high confidence, whose logit margins are closer to correct outputs than to hallucinated ones. Merging them into a single negative class therefore conflates heterogeneous behaviors, introducing inconsistent supervision signals that increase label ambiguity and destabilize training. 

\subsection{Calibration on Downstream Long-Context Tasks}
\label{app:ruler-calibration}

Our out-of-domain experiments demonstrate that the uncertainty detector trained on the synthetic biography data can transfer to RULER without benchmark-specific retraining. We further examine whether a small amount of downstream calibration can improve its context-allocation decisions while preserving this transferable initialization. Specifically, we construct 900 calibration trajectories with QA-4k data split from RULER, using a frozen Llama-3.1-8B-Instruct reader under complete-, partial-, and missing-evidence conditions, targeting \textit{Correct}, \textit{Hallucination}, and \textit{Unknown} responses, respectively. GPT-OSS-120B then assigns token-level labels following the same three-way supervision scheme used for the original uncertainty detector. Starting from the synthetic-pretrained detector, we perform limited calibration on these trajectories and evaluate the resulting detector on the end-to-end RULER tasks.

\begin{table}[t]
\centering
\small
\setlength{\tabcolsep}{3.0pt}
\resizebox{\columnwidth}{!}{
\begin{tabular}{llcccc}
\toprule
& & \multicolumn{2}{c}{\textbf{QA-16k}} 
& \multicolumn{2}{c}{\textbf{QA-8k}} \\
\cmidrule(lr){3-4}
\cmidrule(lr){5-6}
\textbf{Method} & \textbf{Setting} 
& \textbf{mTok} $\downarrow$ & \textbf{Score} $\uparrow$
& \textbf{mTok} $\downarrow$ & \textbf{Score} $\uparrow$ \\
\midrule
Original
& \texttt{topk8/sub4}
& 105.00 & 14.65
& 108.00 & 14.40 \\

Calibration
& \texttt{topk8/sub4}
& \textbf{101.77} & \textbf{19.30}
& \textbf{102.68} & \textbf{23.10} \\
\midrule

Original
& \texttt{topk32/sub16}
& 385.00 & 21.65
& \textbf{352.00} & 26.85 \\

Calibration
& \texttt{topk32/sub16}
& \textbf{364.15} & \textbf{26.10}
& 367.24 & \textbf{26.90} \\
\bottomrule
\end{tabular}
}
\caption{
Effect of limited downstream calibration on RULER. \textsc{Original} denotes the uncertainty detector trained only on the synthetic biography data, while \textsc{Calibration} denotes the same detector after calibration on 900 RULER trajectories. mTok denotes the average number of selected context tokens, and Score follows the official RULER string-match accuracy.
}
\label{tab:ruler-calibration}
\end{table}

As shown in Table~\ref{tab:ruler-calibration}, limited calibration consistently improves generation quality under the compact \texttt{topk8/sub4} setting while also reducing context usage. On QA-16k, the RULER score increases from 14.65 to 19.30 while mTok decreases from 105.00 to 101.77. On QA-8k, calibration yields an even larger improvement from 14.40 to 23.10, together with a reduction in mTok from 108.00 to 102.68. These results indicate that a small amount of task-specific supervision can substantially improve the detector's allocation decisions in the more aggressively compressed regime.
The benefit remains evident with the larger \texttt{topk32/sub16} setting. On QA-16k, calibration improves the score from 21.65 to 26.10 while reducing mTok from 385.00 to 364.15, corresponding to approximately 5.4\% lower context usage. On QA-8k, the score further increases from 26.85 to 26.90, although mTok increases modestly from 352.00 to 367.24. Thus, calibration primarily improves the quality--efficiency trade-off on QA-16k, while preserving the already strong QA-8k performance at the larger context budget.


\subsection{Case Study: Effect of Context Expansion on Token Probability}

To provide a concrete and interpretable analysis of how context window expansion affects the model's predictive confidence, we present a single-generation case study. The example involves extracting the birth date of a person from a long-context passage, a task that exhibits a pronounced uncertainty spike at the day token.

Under the original compact setting ($\texttt{top-k}=1$, $\texttt{block\_size}=16$), the correct token \texttt{``29''} receives a log-probability of $-5.13$ and is ranked third among the candidate tokens \{\texttt{``9''}, \texttt{``15''}, \texttt{``29''}\}. After context window expansion ($\texttt{top-k}=64$), the token \texttt{``29''} becomes the top-ranked candidate, with its log-probability improving to $-4.56$. \tablename~\ref{tab:case_study_for_perplexity_change} summarizes the token rankings and log-probabilities before and after expansion.

\begin{table}[t]
\centering
\fontsize{8.6pt}{10.4pt}\selectfont
\resizebox{\columnwidth}{!}{
\begin{tabular}{lccc}
  \toprule
  \textbf{Setting} & \textbf{Top-1} & \textbf{Rank of \texttt{``29''}} & \textbf{Log $p$(\texttt{``29''}) $\uparrow$} \\
  \midrule
  Before Expansion & \texttt{``9''}  & 3rd & $-5.13$ \\
  After Expansion & \texttt{``29''} & 1st & $-4.56$ \\
  \bottomrule
\end{tabular}
}
\caption{Token ranking and log-probability before and after context expansion. The table reports the candidate ranking for the day token in a birth-date extraction example. Before expansion, the correct token \texttt{``29''} ranks third; after expansion, it becomes top-ranked with log-probability improving from $-5.13$ to $-4.56$.}
~\label{tab:case_study_for_perplexity_change}
\end{table}

This case study illustrates that context expansion can effectively mitigate uncertainty by shifting probability mass toward the correct token. The improvement in both rank and log-probability of \texttt{``29''} after expansion supports the hypothesis that insufficient context leads to degraded token-level confidence, and that restoring broader context allows the model to recover a more discriminative distribution over candidates.

\subsection{Case Study: False-Positive and False-Negative Failure Cases}
\label{app:fp-fn-cases}

To better understand how uncertainty-detection errors affect downstream context allocation, we analyze generation trajectories from the configurations reported in Table~\ref{tab:test-set-exp}. We define a false positive (FP) as an unnecessary context-expansion trigger when the compact context is already sufficient, and a false negative (FN) as a missed trigger when additional evidence is required. We select two representative cases of each error type and perform a single-decision counterfactual replay: for an FP, we suppress one unnecessary expansion; for an FN, we force expansion at one missed trigger. All other inference settings remain unchanged. Because generation is autoregressive, modifying a single detector decision can alter subsequent decoding and therefore the total number of downstream rollback operations.

\begin{table}[t]
\centering
\small
\setlength{\tabcolsep}{3.5pt}
\resizebox{\columnwidth}{!}{
\begin{tabular}{llccc}
\toprule
\textbf{Case} &
\textbf{Intervention} &
\textbf{Quality (\%)} $\uparrow$ &
\textbf{mTokens} $\downarrow$ &
\textbf{Rollbacks} $\downarrow$ \\
\midrule
FP-1 &
Suppress expansion &
100.0 $\rightarrow$ 100.0 &
379.36 $\rightarrow$ 367.83 ($-3.0\%$) &
12 $\rightarrow$ 10 \\

FP-2 &
Suppress expansion &
100.0 $\rightarrow$ 100.0 &
370.14 $\rightarrow$ 359.75 ($-2.8\%$) &
9 $\rightarrow$ 9 \\
\midrule

FN-1 &
Force expansion &
16.7 $\rightarrow$ 66.7 &
173.47 $\rightarrow$ 238.00 ($+37.2\%$) &
1 $\rightarrow$ 6 \\

FN-2 &
Force expansion &
8.3 $\rightarrow$ 75.0 &
316.80 $\rightarrow$ 385.47 ($+21.7\%$) &
1 $\rightarrow$ 9 \\
\bottomrule
\end{tabular}
}
\caption{
Representative false-positive and false-negative failures under single-decision counterfactual replay. We suppress one unnecessary expansion for false positives and force one missed expansion for false negatives. Quality measures per-instance generation quality, mTokens the average attended context tokens, and Rollbacks the total rollback operations.
}
\label{tab:fp-fn-cases}
\end{table}

As shown in Table~\ref{tab:fp-fn-cases}, the two error types have markedly different downstream effects. Correcting false positives preserves generation quality at 100\% while reducing context usage by 2.8--3.0\%, indicating that these errors primarily introduce unnecessary computation. In contrast, correcting false negatives substantially improves per-instance quality, from 16.7\% to 66.7\% in FN-1 and from 8.3\% to 75.0\% in FN-2, at the cost of 37.2\% and 21.7\% more attended context, respectively. The additional context also induces more rollback operations as the corrected trajectory evolves. These cases therefore illustrate an asymmetric failure pattern: false positives mainly reduce context efficiency, whereas false negatives can withhold evidence required for correct generation and consequently have a much larger impact on answer quality. These examples are intended as representative single-trajectory interventions rather than aggregate estimates; the overall quality--efficiency behavior of UT-ACA is reported separately in the main experiments.

\section{Artifact and Data Usage}
\label{sec:artifact-usage}

\paragraph{Pre-trained Models.} We use Llama-3.1-8B-Instruct~\cite{llama3herd2024} and Qwen2-7B-Instruct~\cite{yang2024qwen2} solely for academic research, consistent with respective licenses (Meta Llama 3.1 Community License and Apache 2.0).

\paragraph{Baseline Implementations.} All baseline methods are used strictly for non-commercial academic comparison, consistent with their intended use as research artifacts.

\paragraph{Synthetic Dataset.} Our synthetic biography summarization dataset is constructed entirely in-house and intended for research purposes only. The long-context test set incorporates Wikipedia text retrieved for academic evaluation; Wikipedia content is licensed under CC BY-SA 4.0, and our use is strictly within a non-commercial research context.

\paragraph{Automated Evaluation.} An LLM-as-judge protocol is used for evaluation, consistent with the model's intended research usage terms.